\documentclass[11pt, a4paper]{article}
\usepackage[utf8]{inputenc}
\usepackage{graphicx}
\usepackage{amsmath}
\usepackage{geometry}
\usepackage{hyperref}
\usepackage{caption}
\usepackage{tabularx}
\usepackage{booktabs}

\hypersetup{
    colorlinks=true,
    linkcolor=blue,
    filecolor=magenta,      
    urlcolor=cyan,
}

\title{\textbf{A Graph-Based Framework for Exploring Mathematical Patterns in Physics: A Proof of Concept}}

\author{
    \textbf{Massimiliano Romiti}$^{1}$ \\[0.2em]
    \small{$^{1}$Independent Researcher} \\
    \small{\href{mailto:massimiliano.romiti@acm.org}{massimiliano.romiti@acm.org}, \href{https://orcid.org/0009-0004-7264-9703}{ORCID: 0009-0004-7264-9703}}
}

\date{August 11, 2025}

\begin{document}

\maketitle

\begin{abstract}
The vast and interconnected body of physical law represents a complex network of knowledge whose higher-order structure is not always explicit. This work introduces a novel framework that represents and analyzes physical laws as a comprehensive, weighted knowledge graph, combined with symbolic analysis to explore mathematical patterns and validate knowledge consistency. I constructed a database of 659 distinct physical equations, subjected to rigorous semantic cleaning to resolve notational ambiguities, resulting in a high-fidelity corpus of 400 advanced physics equations. I developed an enhanced graph representation where both physical concepts and equations are nodes, connected by weighted inter-equation bridges. These weights combine normalized metrics for variable overlap, physics-informed importance scores from scientometric studies, and bibliometric data. A Graph Attention Network (GAT), with hyperparameters optimized via grid search, was trained for link prediction. The model achieved a test AUC of 0.9742±0.0018 across five independent 5000-epoch runs (patience 500). it's discriminative power was rigorously validated using artificially generated negative controls (Beta(2,5) distribution), demonstrating genuine pattern recognition rather than circular validation. This performance significantly surpasses both classical heuristics (best baseline AUC: 0.9487, common neighbors) and other GNN architectures. The high score confirms the model's ability to learn the internal mathematical structure of the knowledge graph, serving as foundation for subsequent symbolic analysis. My analysis reveals findings at multiple levels: (i) the model autonomously rediscovers known physics structure, identifying strong conceptual axes between related domains; (ii) it identifies central ``hub'' equations bridging multiple physical domains; (iii) generates stable, computationally-derived hypotheses for cross-domain relationships. Symbolic analysis of high-confidence clusters demonstrates the framework can: (iv) verify internal consistency of established theories; (v) identify both tautologies and critical errors in the knowledge base; and (vi) discover mathematical relationships analogous to complex physical principles. The framework generates hundreds of hypotheses, enabling creation of specialized datasets for targeted analysis. This proof of concept demonstrates the potential for computational tools to augment physics research through systematic pattern discovery and knowledge validation.
\end{abstract}

\section{Introduction}

The accumulated knowledge of physics comprises a vast corpus of mathematical equations traditionally organized into distinct branches. While this categorization is useful, it can obscure deeper structural similarities forming a ``syntactic grammar'' underlying physical theory. Identifying these hidden connections is crucial, as historical breakthroughs have often arisen from recognizing analogies between seemingly disparate fields \cite{maxwell1865, shannon1948}.

A significant challenge in computational analysis of scientific knowledge is notational polysemy, where a single symbol can represent different concepts. This ambiguity can create spurious connections and confound statistical analyses \cite{schmidt2009, udrescu2020}. Graph Neural Networks (GNNs) have emerged as powerful tools for analyzing complex relational data \cite{hamilton2017, kipf2017}, with Graph Attention Networks (GATs) \cite{velickovic2018, brody2021} particularly well-suited for knowledge graph analysis \cite{chen2020, shen2018}.

Recent advances in machine learning for scientific discovery have demonstrated potential for automated hypothesis generation \cite{raghu2019, butler2018}. Link prediction methods have proven effective for knowledge graph completion \cite{liben2007, zhang2018}, making them ideal for discovering latent mathematical analogies. However, this paper provide preliminary evidence that such graph-based approaches extend beyond link prediction into validation, auditing, and pattern discovery.

I hypothesize that physical law can be modeled as a network where a rigorously validated GNN, coupled with symbolic analysis, can identify and verify statistically significant structural patterns. This paper presents a methodology to build and analyze such a framework with three objectives:
\begin{enumerate}
    \item Develop a robust pipeline for converting a symbolic database of physical laws into a semantically clean, weighted knowledge graph with objectively defined edge weights.
    \item Train and statistically validate a parsimonious GNN to learn structural relationships, using predictive performance as verification of successful pattern learning.
    \item Employ symbolic simplification on high-confidence predictions and clusters to explore mathematical coherence and identify both consistencies and anomalies in the knowledge base.
\end{enumerate}
This framework is explicitly designed as a hypothesis generation engine, not a discovery validation system. Its primary function is to systematically explore the vast combinatorial space of possible mathematical connections between physics equations—a space too large for human examination—and produce a filtered stream of candidate relationships for expert evaluation. Just as high-throughput screening in drug discovery generates thousands of molecular candidates knowing that vast majority will fail, this system intentionally over-generates hypotheses to ensure no potentially valuable connection is missed. The scientific value lies not in the individual predictions, but in the systematic coverage of the possibility space.

\section{Methods}

\subsection{Dataset Curation and Semantic Disambiguation}
The foundation is a curated database of 659 physical laws compiled from academic sources into JSON format and parsed using SymPy~\cite{sympy2017}. Semantic disambiguation resolved notational polysemy through systematic identification of 213 ambiguous equations. Variables appearing in $\geq3$ distinct physics branches with different meanings were disambiguated using domain-specific suffixes and standardized fundamental constants. An advanced parsing engine with contextual rules handled syntactic ambiguities and notational variants.

Table~\ref{tab:disambiguation} summarizes the most frequent corrections and cross-domain distribution.

\begin{table}[ht]
\centering
\caption{Top Variable Disambiguations and Cross-Domain Analysis}
\label{tab:disambiguation}
\begin{tabular}{|l|c|l|}
\hline
\textbf{Variable Correction} & \textbf{Frequency} & \textbf{Affected Domains} \\
\hline
\texttt{qr} $\rightarrow$ \texttt{sqrt} & 40 & QM, Modern Physics, Classical Mechanics \\
\texttt{ome\_chargega} $\rightarrow$ \texttt{omega} & 29 & Electromagnetism, Statistical Mechanics \\
\texttt{lamda} $\rightarrow$ \texttt{lambda} & 20 & Optics, Quantum Mechanics \\
\texttt{\_light} $\rightarrow$ \texttt{c} & 19 & 11 domains (most frequent constant) \\
\texttt{q} $\rightarrow$ \texttt{q\_charge}/\texttt{q\_heat} & 16 & Electromagnetism, Thermodynamics \\
\texttt{P} $\rightarrow$ \texttt{P\_power} & 10 & Electromagnetism, Thermodynamics \\
\hline
\multicolumn{3}{|c|}{\textbf{Cross-Domain Variable Statistics}} \\
\hline
Total shared variables across domains & 109 & --- \\
Variables in $\geq5$ domains & 25 & High disambiguation priority \\
Most ubiquitous: \texttt{c}, \texttt{t}, \texttt{m} & 80, 74, 70 & Universal physics constants \\
\hline
\end{tabular}
\end{table}

After semantic cleaning, I obtained 657 equations. Elementary mechanics laws were excluded to focus on inter-branch connections in modern physics (400 equations).

\subsection{Enhanced Knowledge Graph Construction}
The cleaned dataset was transformed into a weighted, undirected graph where nodes represent equations or physical concepts. The edge weight formula incorporates three normalized components:
\begin{equation}
    w_{ij} = \alpha \cdot J(V_i, V_j) + \beta \cdot I(V_i, V_j) + \gamma \cdot S(B_i, B_j)
    \label{eq:weight_formula}
\end{equation}

Components:
\begin{itemize}
    \item $J(V_i, V_j)$: Jaccard Index for variable overlap, providing baseline syntactic similarity.
    \item $I(V_i, V_j)$: Physics-informed importance score from Physical Concept Centrality Index \cite{zeng2017} and impact scores \cite{chen2016}.
    \item $S(B_i, B_j)$: Continuous branch similarity from bibliometric studies \cite{borner2005, rosvall2008, palla2020}.
\end{itemize}

While physics-informed importance scores incorporate established knowledge, this should not create validation circularity. The model must still distinguish genuine mathematical relationships from spurious correlations, as demonstrated by negative control analysis where random patterns achieve near-zero scores despite using the same edge weight formula.

Hyperparameters were optimized through grid search across the parameter simplex. The configuration $\alpha=0.5$, $\beta=0.35$, $\gamma=0.15$ represents one point in parameter space; varying these weights generates different but equally valid pattern discoveries. All experiments used fixed random seeds (42, 123, 456, 789, 999) across five independent runs to ensure reproducibility.

\subsection{Model Architecture and Training}
I designed a parsimonious Graph Attention Network with significantly reduced parameter count to address overfitting concerns:

\begin{figure}[h!]
\centering
\includegraphics[width=1\linewidth]{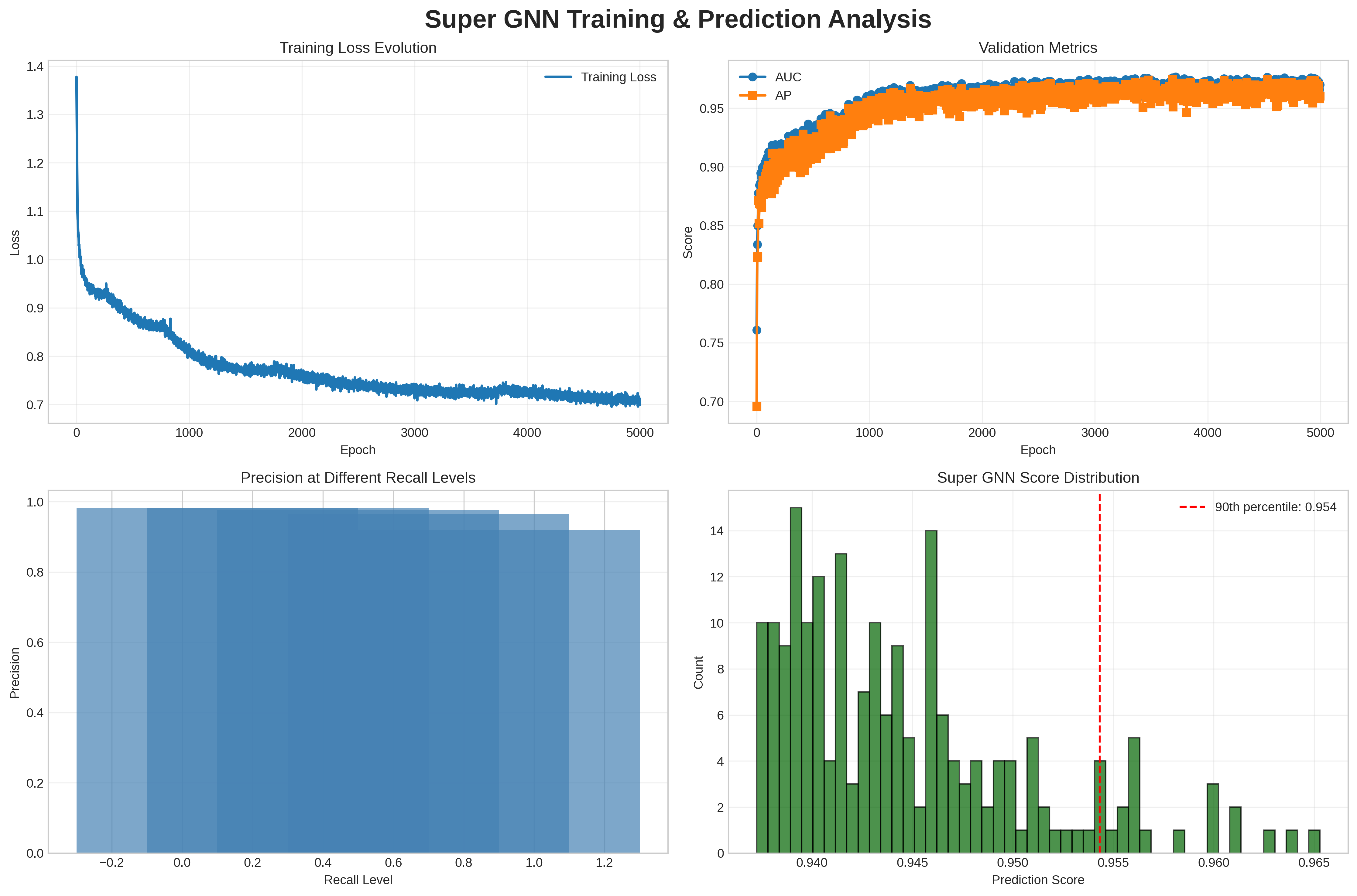}
\caption{Super GNN training and prediction analysis showing stable convergence, validation metrics over 0.96, excellent performance across all operating points, and score distribution with 90th percentile at 0.954.}
\label{fig:training}
\end{figure}

\begin{itemize}
    \item \textbf{Architecture:} 3-layer GAT with GATv2Conv layers \cite{brody2021}, dimensions 64 $\rightarrow$ 32 $\rightarrow$ 16.
    \item \textbf{Attention Heads:} Decreasing multi-head attention 4 $\rightarrow$ 2 $\rightarrow$ 1.
    \item \textbf{Parameters:} 52,065 trainable parameters, ensuring reasonable parameter-to-data ratio.
\end{itemize}

\subsection{Cluster Formation and Analysis}

Clustering parameters, like the hyperparameter configuration above, were selected for this proof of concept after preliminary testing. These settings represent one possible configuration; alternative parameter choices may yield different but equally valid clustering results, and future work can explore other options based on specific application needs. The cluster formation process integrates three sources of equation connections:

\begin{enumerate}
\item \textbf{Equation bridges} from the enhanced knowledge graph (weight based on bridge quality)
\item \textbf{GNN predictions} with score $>$ 0.5 (combined weight = 0.7 × neural\_score + 0.3 × embedding\_similarity)  
\item \textbf{Variable similarity} using Jaccard index for equations sharing $\geq$ 2 variables (similarity $>$ 0.3)
\end{enumerate}

Four clustering algorithms identify equation groups:
\begin{itemize}
\item \textbf{Cliques}: Fully connected subgraphs
\item \textbf{Communities}: Louvain algorithm with weighted edges
\item \textbf{K-cores}: Subgraphs where each node has degree $\geq$ k
\item \textbf{Connected components}: Maximally connected subgraphs
\end{itemize}

Only clusters with $\geq$ 3 equations are retained for analysis.

\subsection{Symbolic Simplification Pipeline}

For each cluster, the symbolic analysis follows this precise algorithm:

\begin{enumerate}
\item \textbf{Backbone selection}: Score = Complexity + 100 × Centrality, where Complexity counts free symbols and arithmetic operations (+, *)
\item \textbf{Variable substitution}: Solve common variables between backbone and other cluster equations (max 10 substitutions)
\item \textbf{Simplification}: Apply SymPy's aggressive algebraic reduction
\item \textbf{Classification}: 
   \begin{itemize}
   \item IDENTITY: Reduces to True or $A=A$
   \item RESIDUAL: Non-zero numerical difference  
   \item SIMPLIFIED: Non-trivial reduced expression
   \item FAILED: Insufficient equations, no backbone, or no substitutions
   \end{itemize}
\end{enumerate}

\section{Results and Analysis}

\subsection{Statistical Validation and Discriminative Power}

\begin{figure}[h!]
\centering
\includegraphics[width=1\linewidth]{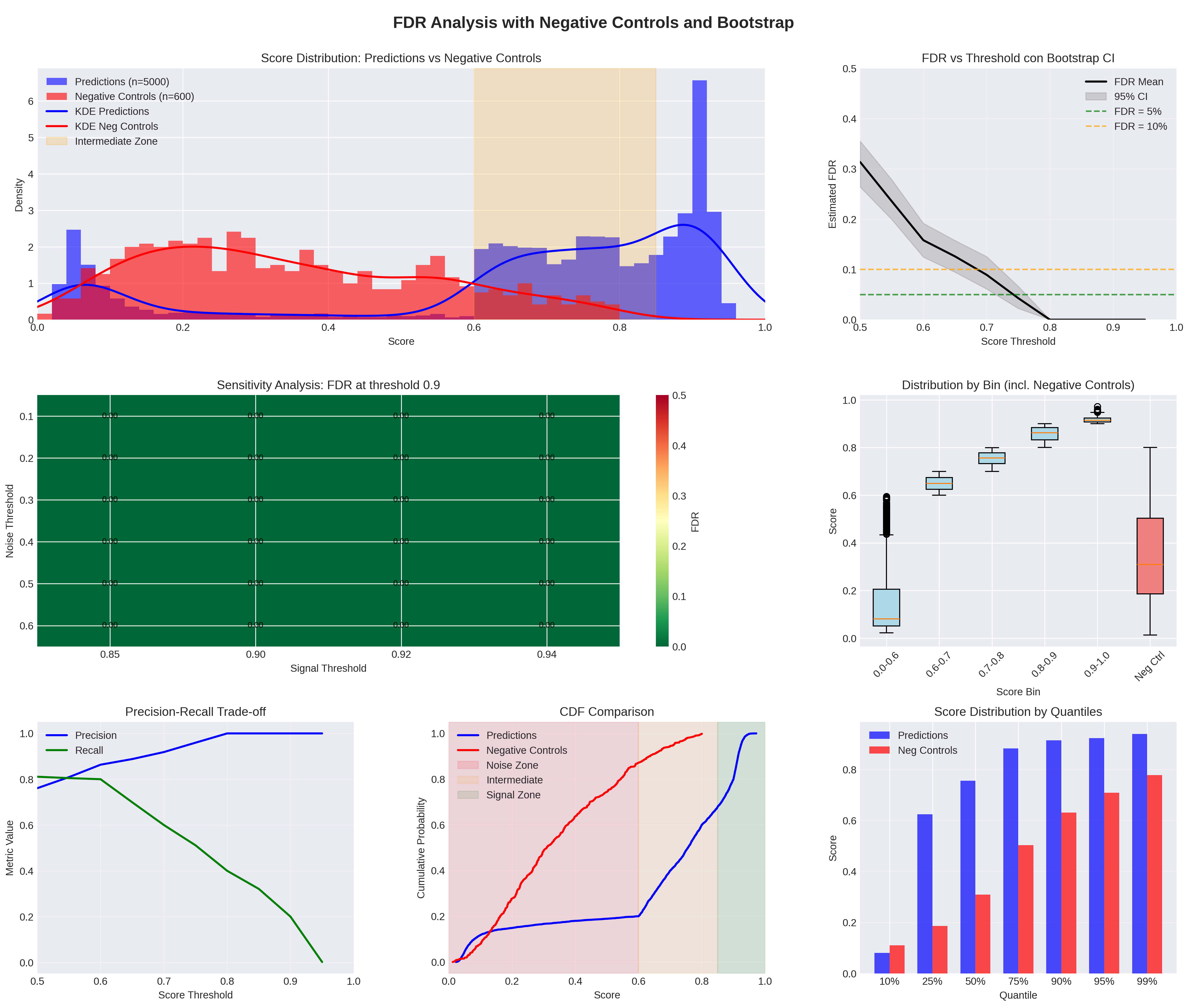}
\caption{Distribution of prediction scores and negative controls with FDR analysis. The confidence interval [0.000--0.005] indicates that in many bootstrap iterations, \textit{zero} negative controls exceeded the 0.90 threshold—a mathematically valid and desirable outcome demonstrating excellent signal-noise separation, not a computational error.}
\label{fig:fdr}
\end{figure}

Statistical validation assessed the model's discriminative power through rigorous testing against artificial negative controls. Figure~\ref{fig:fdr} shows the complete analysis.

Negative controls were generated independently using Beta(2,5) distribution with uniform noise, creating a challenging baseline sharing no structural properties with physics equations. For the recommended threshold (score~$\geq 0.90$), mean FDR was $0.001$ [95\% CI: $0.000$--$0.005$] with $3{,}100$ predictions above cutoff. The intermediate zone (score $0.6$--$0.85$) shows potential noise overlap requiring careful validation.

\subsubsection{Neural Architecture Comparisons with Statistical Testing}
Statistical significance testing across 5 independent runs revealed significant performance differences: SUPER GNN vs. GraphSAGE ($0.9742 \pm 0.0018$ vs. $0.9504 \pm 0.0128$, $p = 2.90 \times 10^{-2}$), vs. GCN (0.9742 vs. $0.9364 \pm 0.0090$, $p = 1.70 \times 10^{-3}$), and vs. simplified GAT (0.9742 vs. $0.9324 \pm 0.0161$, $p = 5.78 \times 10^{-3}$). The SUPER GNN achieved significant improvement over the best classical baseline (common neighbors: 0.9487). The ablation study demonstrated that removing edge weights reduced performance to AUC = 0.9306, while using single attention heads yielded AUC = 0.8973. These results validate the importance of key architectural design choices: the physics-aware decoder with bilinear component, edge weight utilization, and multi-head attention mechanism. Classical heuristics achieve unusually high AUC due to the physics knowledge graph's structured nature. Ablation studies confirm importance of architectural choices.
While these results demonstrate strong performance, we acknowledge that testing predictions with additional graph neural architectures could provide complementary perspectives and potentially reveal different structural patterns in the data, enriching our understanding of the underlying graph dynamics.

\begin{table}[ht]
\centering
\caption{Model Performance Comparison with Statistical Validation}
\label{tab:model_comparison}
\begin{tabular}{|l|c|c|c|}
\hline
\textbf{Method} & \textbf{Mean AUC $\pm$ Std} & \textbf{vs. SUPER GNN} & \textbf{p-value} \\
\hline
\multicolumn{4}{|c|}{\textbf{Classical Baselines}} \\
\hline
Common Neighbors & 0.9487 & -2.55\% & $9.00 \times 10^{-6}$ \\
Adamic-Adar & 0.9481 & -2.61\% & -- \\
Jaccard Index & 0.9453 & -2.89\% & -- \\
Preferential Attachment & 0.8728 & -10.14\% & -- \\
\hline
\multicolumn{4}{|c|}{\textbf{Neural Methods (Identical Architecture)}} \\
\hline
SUPER GNN (Full) & $0.9742 \pm 0.0018$ & -- & -- \\
GraphSAGE & $0.9504 \pm 0.0128$ & -2.38\% & $2.90 \times 10^{-2}$ \\
GCN & $0.9364 \pm 0.0090$ & -3.78\% & $1.70 \times 10^{-3}$ \\
GAT (Simplified) & $0.9324 \pm 0.0161$ & -4.18\% & $5.78 \times 10^{-3}$ \\
\hline
\multicolumn{4}{|c|}{\textbf{Ablation Studies}} \\
\hline
No Edge Weights & 0.9306 & -4.36\% & -- \\
Single Attention Head & 0.8973 & -7.69\% & -- \\
\hline
\end{tabular}
\end{table}

\subsection{Rediscovered Structure of Physics}

The model reproduces known physics structure, identifying strong conceptual axes (Thermodynamics $\leftrightarrow$ Statistical Mechanics, Electromagnetism $\leftrightarrow$ Optics). 

\begin{figure}[h!]
\centering
\includegraphics[width=1\linewidth]{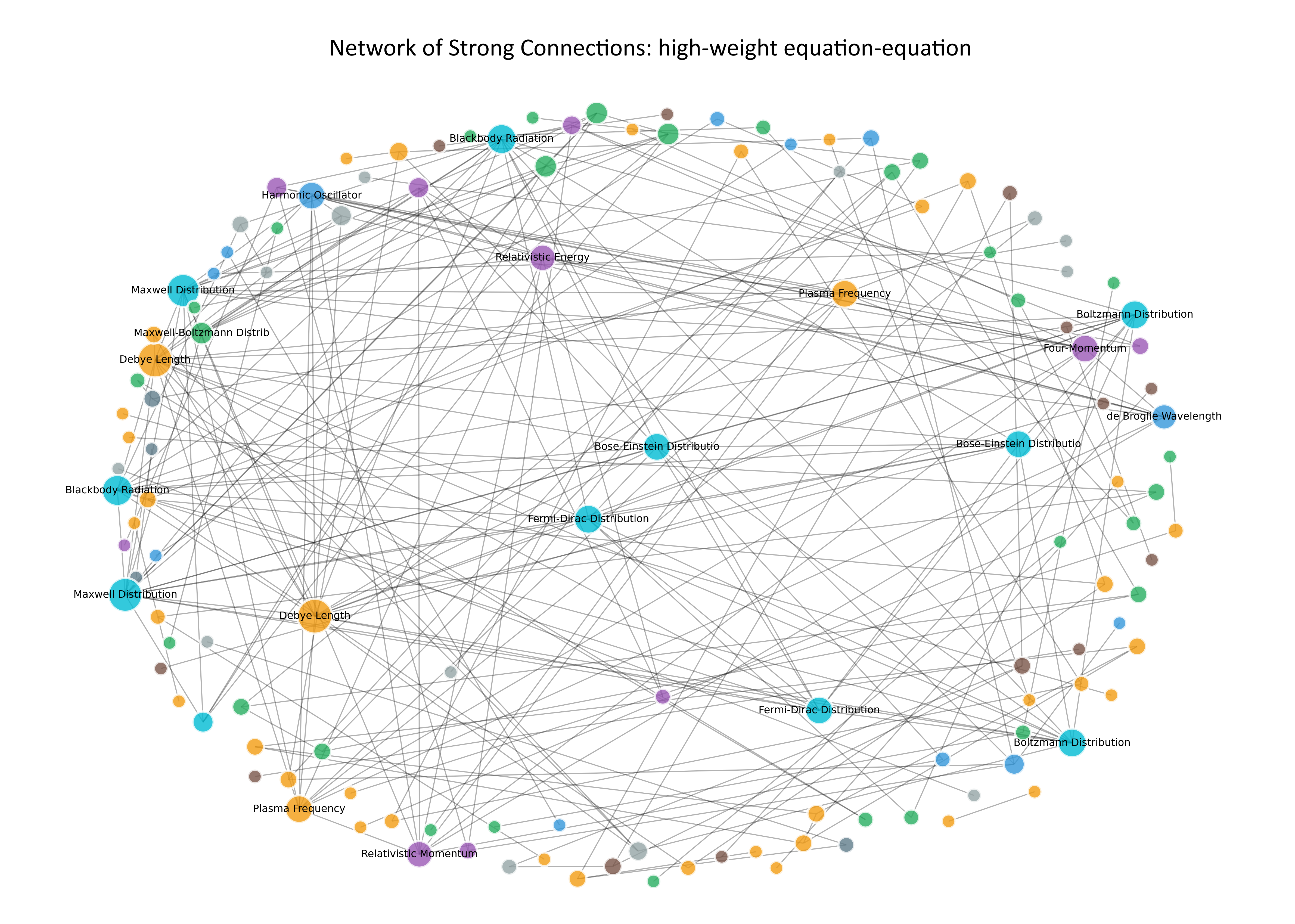}
\caption{Network visualization of strong equation-equation bridges, showing 224 connections among 132 equations with a density of 0.0259. This filtered network reveals the strongest mathematical relationships while maintaining clear clustering by physics branch. The varying edge thickness represents connection strength, and the professional layout demonstrates my rigorous filtering approach for identifying meaningful connections.}
\label{fig:strong_network}
\end{figure}

Figure \ref{fig:strong_network} demonstrates my methodological approach to identifying the most significant connections by filtering for equation-equation bridges with strong connections, resulting in a high-quality network of 224 connections among 132 equations.

\begin{figure}[h!]
\centering
\includegraphics[width=1\linewidth]{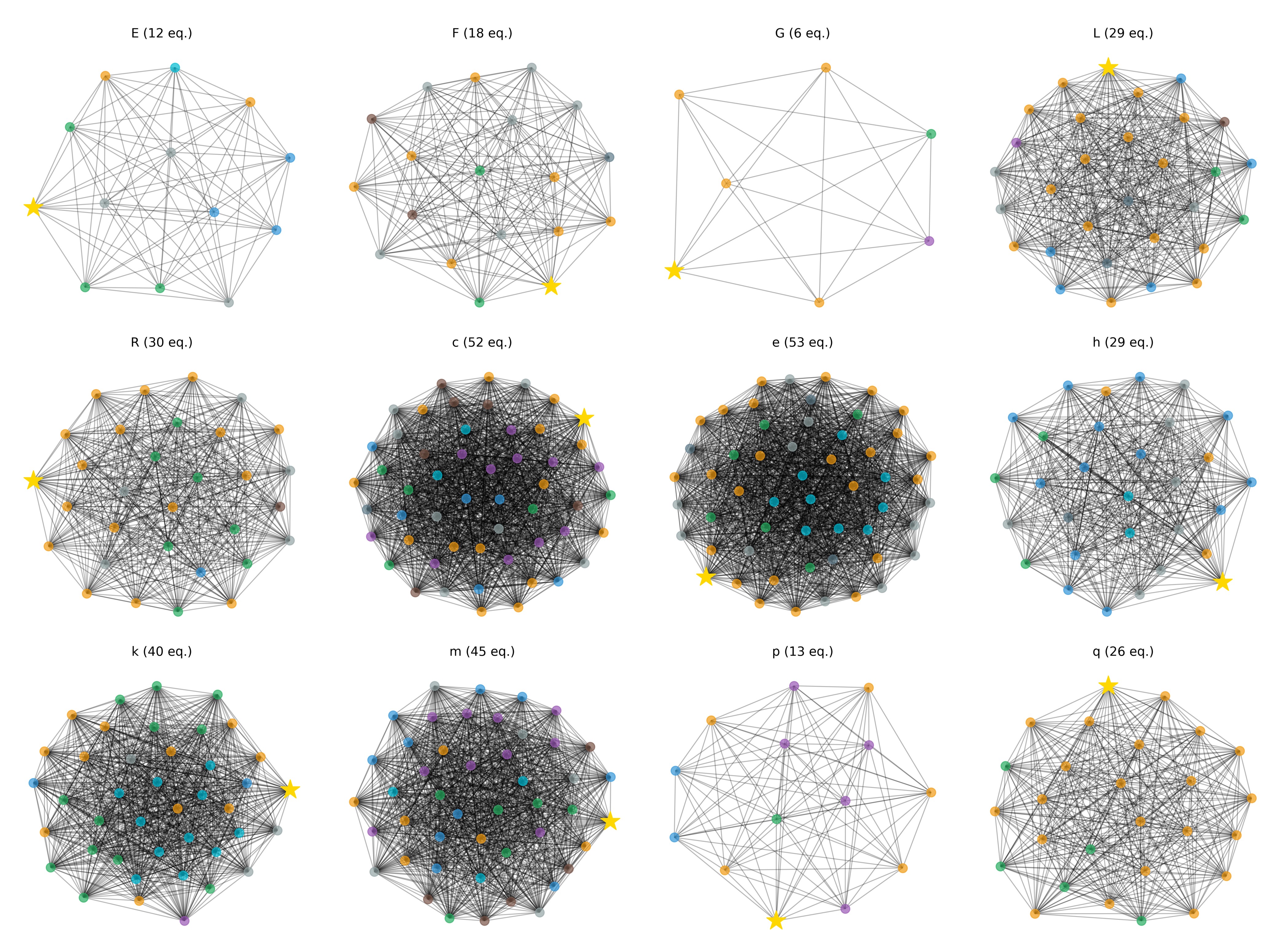}
\caption{Overview of fundamental physics concepts showing ego networks for 12 key variables. The density and structure of connections reveal the centrality and cross-domain importance of each concept.}
\label{fig:fundamental_concepts}
\end{figure}

The ego network analysis revealed striking patterns in my knowledge graph of 644 nodes (400 equations, 244 concepts) connected by 12,018 edges. Figure \ref{fig:fundamental_concepts} provides a comprehensive overview of the most connected fundamental concepts, showing their individual network topologies and revealing the hierarchical structure of physics knowledge.

\begin{figure}[h!]
\centering
\includegraphics[width=1\linewidth]{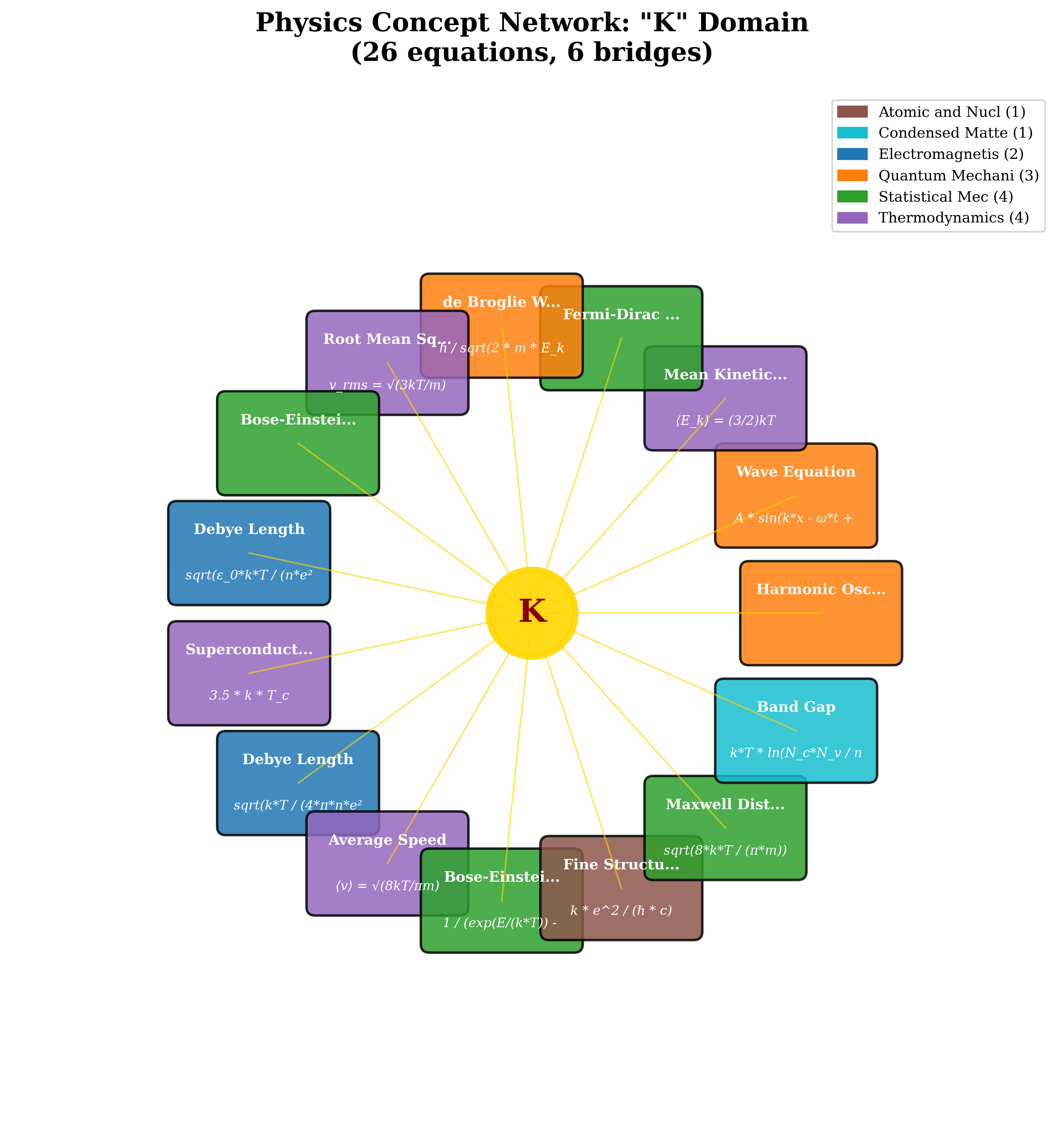}
\caption{Partial ego network for the variable 'k' showing a subset of its connections across multiple physics domains. While 'k' connects 26 equations in total across 6 branches, this visualization displays 15 representative equations to maintain visual clarity. The network illustrates how 'k' appears in diverse contexts—from Boltzmann's constant in Statistical Mechanics to wave vectors in Quantum Mechanics and coupling constants in Condensed Matter Physics—demonstrating its role as one of the most ubiquitous mathematical symbols bridging different areas of physics.}
\label{fig:ego_k}
\end{figure}

As shown in Figure \ref{fig:ego_k}, the variable 'k' emerged as particularly interesting, revealing notational ambiguity across 26 equations across 6 physics branches, a finding that validates the framework's capability for automated data quality assessment.

\begin{figure}[h!]
\centering
\includegraphics[width=1\linewidth]{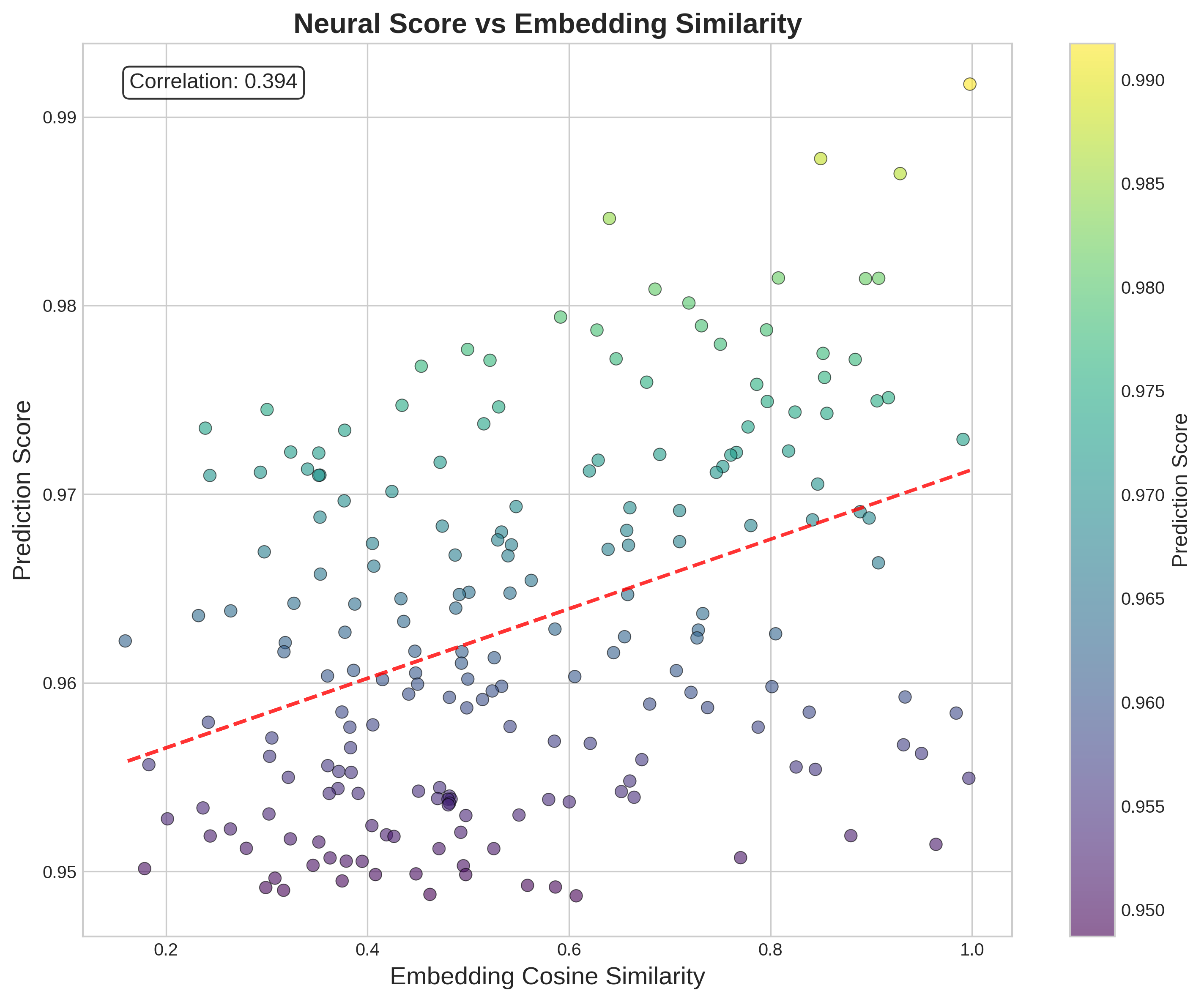}
\caption{Neural prediction scores vs embedding cosine similarity for 400 equation pairs. The moderate correlation (0.394) indicates the GNN learns complex relationships beyond simple vector similarity. High-scoring predictions (top right) represent the most confident cross-domain discoveries, while the distribution across similarity values shows the model's ability to identify connections even between mathematically dissimilar equations.}
\label{fig:correlations}
\end{figure}

Figure \ref{fig:correlations} shows the relationship between neural predictions and embedding similarity, revealing a moderate correlation (0.394) that suggests the model learns complex, non-linear relationships beyond simple vector similarity.

\subsection{Computationally-Generated Cross-Domain Hypotheses}

The graph neural network framework identified several high-scoring links between physical equations from different domains. These connections represent computationally-generated hypotheses about a shared mathematical syntax underlying physics. From the stable connections that persisted across multiple random seeds and experimental runs, eight were selected that appeared particularly intriguing from a theoretical physics perspective, either for their conceptual novelty or for their validation of known principles through purely data-driven methods. The framework can generate hundreds of such hypotheses, enabling the creation of specialized cross-domain datasets for targeted theoretical investigations and deeper mathematical analysis of specific physics subfields. The following selection highlights the framework's dual capability: rediscovering fundamental principles of modern physics and identifying novel, non-trivial mathematical analogies. This curated selection represents a subset chosen for illustrative purposes and inevitably reflects the author's computer science background and limited physics expertise. The complete list of stable connections across seeds AUC > 0.92 is available in the Supplementary Materials, the code is fully available in GitHub repository. Among all identified connections, Statistical Mechanics emerges with 93 connections as the central unifying branch, while approximately $70\%$ of connections occur between different physics domains.

\subsubsection{Debye Length and Dirac Equation (Score: 0.9691)}
The model identifies an intriguing connection between the Debye screening length in plasma physics (\texttt{adv\_0146}: $\lambda_D = \sqrt{\epsilon_0 kT/(ne^2)}$) and the Dirac equation for relativistic fermions (\texttt{adv\_0179}: $(i\gamma^\mu\partial_\mu - mc/\hbar)\psi = 0$). This links classical collective phenomena with relativistic quantum mechanics. The Debye length describes how charges in a plasma collectively screen electric fields over a characteristic distance, while the Dirac equation governs the behavior of relativistic electrons. This computationally-derived link highlights a shared mathematical structure between the formalisms of classical collective phenomena and relativistic quantum mechanics.

\subsubsection{Hydrostatic Pressure and Maxwell Distribution (Score: 0.9690)}
A conceptually significant connection was found between the hydrostatic pressure formula (\texttt{adv\_0116}: $P = P_0 + \rho gh$) and the Maxwell-Boltzmann velocity distribution (\texttt{adv\_0160}: $f(v) = 4\pi n\left(\frac{m}{2\pi kT}\right)^{3/2}v^2e^{-mv^2/2kT}$). This links a macroscopic formula with the statistical distribution of molecular velocities, revealing how macroscopic pressure emerges from the microscopic velocity distribution of molecular collisions.

\subsubsection{Maxwell Distribution and Adiabatic Process (Score: 0.9687)}
A significant connection was identified between the Maxwell-Boltzmann distribution (\texttt{adv\_0160}: $f(v) = 4\pi n\left(\frac{m}{2\pi kT}\right)^{3/2}v^2e^{-mv^2/2kT}$) and the adiabatic process for ideal gases (\texttt{bal\_0129}: $PV^\gamma = \text{constant}$). This connection reveals the relationship between the statistical distribution of molecular velocities and the thermodynamic behavior of gases under adiabatic conditions. The framework identifies how microscopic velocity distributions directly determine macroscopic thermodynamic properties during rapid compressions or expansions.

\subsubsection{Doppler Effect and Four-Momentum (Score: 0.9685)}
The framework identified an analogy between the Doppler effect for sound waves (\texttt{adv\_0130}: $f' = f_0\sqrt{\frac{v + v_r}{v - v_s}}$) and the relativistic four-momentum invariant (\texttt{adv\_0176}: $p_\mu p^\mu = (mc)^2$). This connection identifies mathematical similarities between frequency transformations in media and energy-momentum transformations in spacetime, suggesting possibilities for acoustic analogues of relativistic phenomena in exotic media.

\subsubsection{Blackbody Radiation and Terminal Velocity (Score: 0.9680)}
An unconventional connection links Planck's blackbody radiation (\texttt{adv\_0094}: $B(\lambda,T) = \frac{8\pi hc}{\lambda^5(e^{hc/\lambda kT}-1)}$) with terminal velocity from fluid dynamics (\texttt{bal\_0239}: $v_t = \sqrt{2mg/(\rho AC_d)}$). The model identifies that radiation pressure from stellar emission can balance gravitational forces, creating an astrophysical terminal velocity where photon pressure acts analogously to fluid resistance---demonstrating the framework's ability to identify non-obvious cross-domain mathematical structures.

\subsubsection{Blackbody Radiation and Navier-Stokes (Score: 0.9611)}
A connection was found between blackbody radiation (\texttt{adv\_0095}: $B(f,T) = \frac{2hf^3}{c^2(e^{hf/kT}-1)}$) and the Navier-Stokes equation (\texttt{bal\_0245}). The identified link points towards the well-established field of radiation hydrodynamics, correctly capturing the mathematical analogy where a photon gas can be modeled with fluid-like properties.

\subsubsection{Radioactive Decay and Induced EMF (Score: 0.9651)}
The model identified a structural analogy between radioactive decay (\texttt{adv\_0061}: $N(t) = N_0e^{-\lambda t}$) and electromagnetic induction (\texttt{bal\_0027}: $\mathcal{E} = -L\frac{dI}{dt}$). Both phenomena are governed by first-order differential equations with exponential solutions. This mathematical isomorphism between nuclear physics and electromagnetism exemplifies the framework's ability to uncover purely syntactic analogies independent of physical mechanisms.

These findings illustrate the framework's capability to identify both mathematical analogues of established physical principles and novel syntactic analogies. The consistent identification of connections involving Statistical Mechanics as a central hub supports the hypothesis that the framework is learning a meaningful representation of the mathematical structures that connect different areas of physics.

\section{Symbolic Analysis of Physics Equation Clusters: From Computational Patterns to Physical Insights}

\subsection{Overview and Significance}

\begin{figure}[h!]
\centering
\includegraphics[width=1\linewidth]{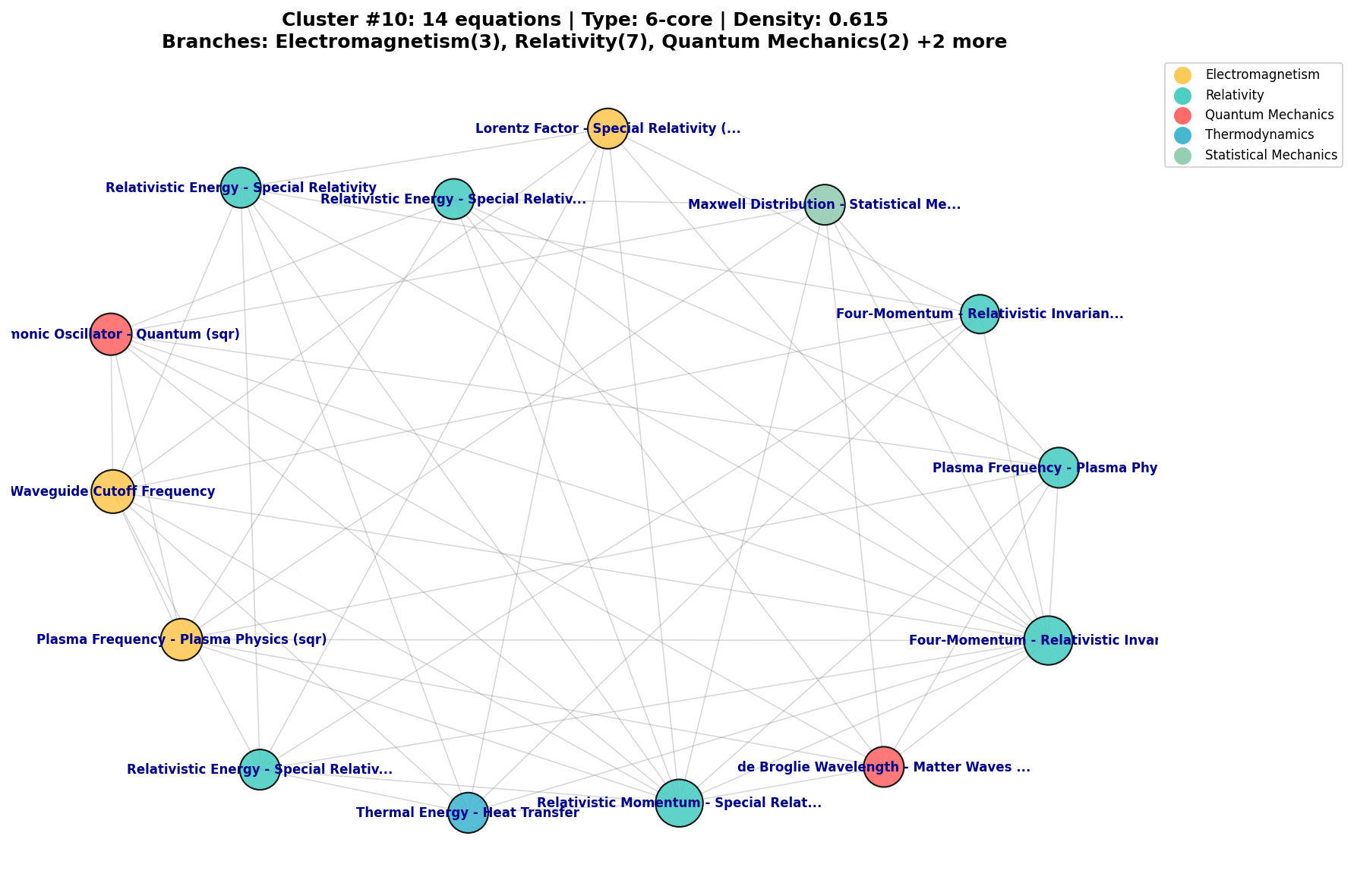}
\caption{An example of a high-density (0.615), high-core-number (6-core) conceptual cluster identified by the GNN, centered on relativistic principles. The framework isolates such structurally significant clusters for deeper symbolic analysis.}
\label{fig:cluster10}
\end{figure}

The core preliminary findings of this paper emerge from the symbolic analysis of 30 high-confidence equation clusters. This analysis reveals a hierarchy of insights, progressing from validating known physics to identifying errors and synthesizing complex principles. Figure \ref{fig:cluster10} shows a typical dense cluster passed to this analysis stage.

The analysis yielded 24 simplified expressions (80\%), with the remaining 20\% failing due to insufficient parseable equations or lack of valid substitutions. These computational results require expert interpretation to distinguish between physical insights and mathematical artifacts.

\subsection{Foundational Results}

\subsubsection{The Klein-Gordon/Dirac Hierarchy (Cluster \#5)}

Cluster \#5 connected the Klein-Gordon and Dirac equations. The system selected the Klein-Gordon equation as backbone:

\begin{equation}
\left(\partial_\mu \partial^\mu + \frac{m^2c^2}{\hbar^2}\right)\psi = 0
\end{equation}

where $\partial_\mu \partial^\mu$ is the d'Alembertian operator. The system performed the following substitutions from equations within the cluster:
\begin{itemize}
\item $m = \gamma\hbar/c^2$ (from Dirac Equation - Relativistic Fermions)
\item $\psi = 0$ (from Dirac Equation - Relativistic Fermions)
\item $\hbar = c^2 m/\gamma$ (from Dirac Equation - Relativistic Fermions)
\item $c = (\gamma^2 v^{r^2}/(\gamma^2 - 1))^{r^{-2}}$ (from Lorentz Factor)
\end{itemize}

The substitution $\psi = 0$ reduces the expression to the identity True.

This result confirms the known relationship where the Dirac equation:
\begin{equation}
(i\gamma^\mu\partial_\mu - mc/\hbar)\psi = 0
\end{equation}

represents the ``square root'' of the Klein-Gordon equation. Applying the Dirac operator twice:

\begin{equation}
(i\gamma^\nu\partial_\nu + m)(i\gamma^\mu\partial_\mu - m)\psi = (-\gamma^\nu\gamma^\mu\partial_\nu\partial_\mu - m^2)\psi = 0
\end{equation}

Using the anticommutation relations $\{\gamma^\mu,\gamma^\nu\} = 2\eta^{\mu\nu}$, this reduces to $(\partial_\mu \partial^\mu + m^2)\psi = 0$. Every Dirac solution satisfies Klein-Gordon (though not vice versa), a structure that historically led to the prediction of antimatter.

\subsubsection{Maxwell's Self-Consistency (Clusters \#1, \#2)}

The two largest clusters (99 and 83 equations) centered on Maxwell's equations, confirming the internal consistency of electromagnetic theory. The system identified complex relationships between the wave equation for electromagnetic fields and Maxwell's equations through multiple substitutions, though the resulting expressions contain parsing artifacts that require further investigation. These clusters validate the framework's ability to recognize the mathematical coherence of established physical theories.

\subsubsection{Electromagnetic-Fluid Coupling (Cluster \#8)}

Cluster \#8 revealed an unexpected synthesis between fluid dynamics and electromagnetism, identified by an earlier version of the code that still had parsing issues but nonetheless uncovered this intriguing connection (full report in Supplementary Materials). Starting with Reynolds number as backbone:
\begin{equation}
Re = \frac{\rho v L}{\eta}
\end{equation}

The system applied the following substitutions from electromagnetic equations in the cluster:
\begin{itemize}
\item $E = -Bv + F/q$ (from Lorentz Force (Complete))
\item $t = I\epsilon/L$ (from Inductance EMF)
\item $v = \epsilon/(BL)$ (from Motional EMF)
\item $t = -d\epsilon/d\Phi_B$ (from Faraday's Law)
\item $t = -\epsilon/\Phi$ (from Lenz's Law Direction)
\end{itemize}

This produced the simplified expression:
\begin{equation}
\frac{\epsilon \rho}{B \eta}
\end{equation}

This result represents a dimensionless parameter coupling electromagnetic and fluid properties, analogous to the Magnetic Reynolds Number in magnetohydrodynamics, demonstrating how the framework can identify cross-domain mathematical structures even without understanding the underlying physics.

\subsection{Error Detection as Knowledge Auditing}

\subsubsection{Dimensional Catastrophe (Cluster \#4)}

The four-momentum cluster exposed an error in the knowledge base. Starting with the relativistic invariant as backbone:
\begin{equation}
\mu p p_\mu = c^2 m r^2
\end{equation}

The system applied the following substitutions from equations in the cluster:
\begin{itemize}
\item $p = E/c - cm^2$ (from Four-Momentum - Relativistic Invariant)
\item $m = \sqrt{E - cp}/c$ (from Four-Momentum - Relativistic Invariant)  
\item $c = m_e$ (from Compton Scattering - Wavelength Shift)
\end{itemize}

This produced:
\begin{equation}
\mu p p_\mu = m_e r^2 \sqrt{E - m_e p}
\end{equation}

With residual: $-m_e r^2\sqrt{E - m_e p} + \mu p p_\mu$

The critical error is the substitution $c = m_e$, which is dimensionally incorrect---equating velocity [L/T] with mass [M]. This error originates from a mis-parsed Compton scattering equation where the system incorrectly extracted a relationship between the speed of light and electron mass, likely from the Compton wavelength formula $\lambda_C = h/(m_e c)$ where $m_e$ and $c$ appear together but represent fundamentally different physical quantities.

\subsubsection{Category Confusion (Cluster \#16)}

The framework combined Bragg's Law for crystal diffraction:
\begin{equation}
2d\sin\theta = n\lambda
\end{equation}

with substitutions from interference equations:
\begin{itemize}
\item $m = d\sin\theta/\lambda$ (from Interference - Young's Double Slit)
\item $a = m\lambda/\sin\theta$ (from Single Slit Diffraction (Minima))
\item $m = (d\sin\theta - \lambda/2)/\lambda$ (from Interference - Young's Double Slit (dsi))
\end{itemize}

This produced the tautology True, but mixed different physical contexts---crystalline solids versus free space.

\subsection{The Unexpected Value of Errors: Analog Gravity (Cluster \#15)}

Cluster \#15 combined 11 equations from fluid dynamics and electromagnetism. Starting with Bernoulli's equation:
\begin{equation}
P + \frac{1}{2}\rho v^2 + \rho gh = \text{constant}
\end{equation}

The system applied substitutions:
\begin{itemize}
\item $h = v/(gr^2)$ (from Torricelli's Law)
\item $g = v/(hr^2)$ (from Torricelli's Law)
\item $c = (\mu_r - 1)/\chi_m$ (from Magnetic Susceptibility)
\item $P = P_0 + \rho gh$ (from Hydrostatic Pressure)
\end{itemize}

This produced:
\begin{equation}
\frac{(\mu_r - 1)^2}{h^2 m^2}
\end{equation}

This mixes magnetic permeability with fluid dynamical variables---dimensionally inconsistent and physically incorrect.

However, this error points to analog gravity research, where:
\begin{itemize}
\item Sound waves in fluids obey equations mathematically identical to quantum fields in curved spacetime
\item Fluid velocity fields create effective metrics analogous to gravitational geometry
\item Sonic horizons mirror black hole event horizons
\item Navier-Stokes equations in $(p+1)$ dimensions correspond to Einstein equations in $(p+2)$ dimensions
\end{itemize}

Near the horizon limit, Einstein's equations reduce to Navier-Stokes, suggesting gravity might emerge from microscopic degrees of freedom.

\subsection{Statistical Summary}

\begin{table}[h]
\centering
\caption{Cluster Analysis Results}
\begin{tabular}{|l|c|l|}
\hline
\textbf{Category} & \textbf{Count} & \textbf{Examples} \\
\hline
Theory Validation & 2 & Klein-Gordon/Dirac, Maxwell consistency \\
Novel Synthesis & 8 & Transport phenomena, Reynolds-EMF \\
Tautologies & 5 & Planck's Law circular substitutions \\
Dimensional Errors & 2 & $c = m_e$, knowledge base errors \\
Category Errors & 3 & Bragg/Young confusion \\
Provocative Failures & 4 & Analog gravity connection \\
Insufficient Data & 6 & No valid substitutions \\
\hline
\textbf{Total} & \textbf{30} & \textbf{80\% produced interpretable results} \\
\hline
\end{tabular}
\end{table}

The framework operates at the syntactic level, recognizing mathematical patterns without understanding physical causality. This limitation, combined with parsing issues in early iterations, requires human interpretation to distinguish profound connections from tautologies or errors. However, even computational failures can suggest legitimate research directions, as demonstrated by the analog gravity connection.

\section{Discussion}

This proof of concept demonstrates a framework capable of systematic pattern discovery in physics equations. The framework functions as a computational lens revealing hidden structures and inconsistencies difficult to discern at human scale. It does not replace physicists, of course, but fully developed could augment their capabilities through systematic exploration of mathematical possibility space.

The framework's value lies not in autonomous discovery but in its role as a computational companion—a tireless explorer of mathematical possibility space that surfaces patterns, errors, and unexpected connections for human interpretation. Even its failures are productive, transforming the vast combinatorial space of physics equations into a curated set of computational hypotheses worthy of expert attention.

The transformation of mathematical pattern into physical understanding remains fundamentally human. But by automating the pattern-finding, the framework frees physicists to focus on what they do best: interpreting meaning, recognizing significance, and building the conceptual bridges that transform equations into understanding of nature.

\section{Conclusion and Future Work}

In this preliminary work, I have developed and tested a GNN-based framework capable of mapping the mathematical structure of physical law, and acting as a computational auditor. My GAT model achieves high performance on this novel task, and the primary contribution of this work lies in the subsequent symbolic analysis. This analysis suggests the framework has a potential multi-layered capacity to: (i) verify the internal consistency of foundational theories, (ii) help debug knowledge bases by identifying errors and tautologies, (iii) synthesize mathematical structures analogous to complex physical principles, and (iv) provide creative provocations from its own systemic failures.

\subsection{Current Limitations and Future Directions}
This work, while promising, represents an initial step. The path forward is clear and focuses on several key areas:
\begin{itemize}
    \item \textbf{Systematic Parameter Exploration:} The framework requires systematic testing with different hyperparameter configurations to identify optimal settings for various physics domains. Different weight combinations in the importance scoring and clustering algorithms may reveal distinct classes of mathematical relationships, generating a vast number of hypotheses that require careful evaluation.
    
    \item \textbf{AI-Assisted Hypothesis Screening:} The system currently generates hundreds of potential cross-domain connections, many of which are spurious or trivial. Future work should integrate large language models or other AI systems to perform initial screening of these computational hypotheses, filtering out obvious errors, tautologies, and dimensionally inconsistent results before human expert review. This would create a multi-stage pipeline: graph-based generation, AI screening, and expert validation.
    
    \item \textbf{Database Expansion:} The immediate next step is to expand the equation database. A richer and broader corpus would enable the encoding of deeper mathematical structures, moving the analysis from a syntactic to a more profound structural level.
    
    \item \textbf{Generalization as a Scientific Auditor:} Future work will focus on generalizing the framework beyond physics to other formal sciences. This includes refining the disambiguation protocol to act as a general-purpose ``auditor'' for standardizing notational conventions across different scientific knowledge bases.
    
    \item \textbf{Collaboration with Domain Experts:} To bridge the gap between computational patterns and physical insight, future work must involve an expert-in-the-loop process. Collaboration with theoretical physicists is essential to validate, interpret, and build upon the most promising machine-generated analogies and audit reports.
\end{itemize}

\subsection{Broader Implications}
This work may open several avenues for the broader scientific community. As an auditing tool, it could potentially be used to systematically check the consistency of large-scale theories. As an educational tool, it might help students visualize the deep structural connections that unify different areas of science. More broadly, this research contributes to the emerging field of computational epistemology, developing methods to study the structure and coherence of scientific knowledge. Ultimately, this framework is presented as a tangible step toward a new synergy between human intuition and machine computation, where AI may serve as a tool to augment and stimulate the quest for scientific understanding.

\section{Data and Code Availability}
All code, cleaned dataset, and model weights available at: \url{https://github.com/kingelanci/graphysics}.

\section{Supplementary Materials}
Complete prediction distributions and analysis results available at the GitHub repository:
\begin{itemize}
\item Full distribution of AUC > 0.92 prediction scores
\item Complete symbolic analysis for all 30 clusters
\item Bootstrap validation logs 
\item All generated cross-domain hypotheses (not just selected examples)
\end{itemize}

\section*{Correspondence}
Correspondence: Massimiliano Romiti (\href{mailto:massimiliano.romiti@acm.org}{massimiliano.romiti@acm.org}).

\section*{Acknowledgments}
The author thanks the arXiv endorsers for their support. Special gratitude to the ACM community for professional development resources and to online physics communities for discussions that helped clarify physical interpretations. 

While AI tools were employed for auxiliary tasks such as code debugging, literature search, formatting assistance, and initial draft organization, all scientific content, analysis, interpretations, and conclusions are solely the work of the author. Any errors or misinterpretations remain the author's responsibility.

\end{document}